\documentclass[conference]{IEEEtran}
\IEEEoverridecommandlockouts

\usepackage{amsmath,amssymb,amsfonts}
\usepackage{algorithmic}
\usepackage{graphicx}
\usepackage{textcomp}
\usepackage{xcolor}
\usepackage{hyperref}
\usepackage{tabularx,multirow}
\usepackage{booktabs}
\usepackage{adjustbox}
\usepackage[numbers,sort&compress]{natbib}

\def\BibTeX{{\rm B\kern-.05em{\sc i\kern-.025em b}\kern-.08em
    T\kern-.1667em\lower.7ex\hbox{E}\kern-.125emX}}

\begin{document}

\title{Evaluating Team Skill Aggregation in \\ Online Competitive Games}


\author{\IEEEauthorblockN{Arman Dehpanah}
\IEEEauthorblockA{\textit{School of Computing} \\
\textit{DePaul University}\\
Chicago, USA \\
adehpana@depaul.edu}
\and
\IEEEauthorblockN{Muheeb Faizan Ghori}
\IEEEauthorblockA{\textit{School of Computing} \\
\textit{DePaul University}\\
Chicago, USA \\
mghori2@depaul.edu}
\and
\IEEEauthorblockN{Jonathan Gemmell}
\IEEEauthorblockA{\textit{School of Computing} \\
\textit{DePaul University}\\
Chicago, USA \\
jgemmell@cdm.depaul.edu}
\and
\IEEEauthorblockN{Bamshad Mobasher}
\IEEEauthorblockA{\textit{School of Computing} \\
\textit{DePaul University}\\
Chicago, USA \\
mobasher@cs.depaul.edu}}

\maketitle


\begin{abstract}

One of the main goals of online competitive games is increasing player engagement by ensuring fair matches.
These games use rating systems for creating balanced match-ups.
Rating systems leverage statistical estimation to rate players' skills and use skill ratings to predict rank before matching players.
Skill ratings of individual players can be aggregated to compute the skill level of a team.
While research often aims to improve the accuracy of skill estimation and fairness of match-ups, less attention has been given to how the skill level of a team is calculated from the skill level of its members.
In this paper, we propose two new aggregation methods and compare them with a standard approach extensively used in the research literature.
We present an exhaustive analysis of the impact of these methods on the predictive performance of rating systems.
We perform our experiments using three popular rating systems, Elo, Glicko, and TrueSkill, on three real-world datasets including over 100,000 battle royale and head-to-head matches.
Our evaluations show the superiority of the MAX method over the other two methods in the majority of the tested cases, implying that the overall performance of a team is best determined by the performance of its most skilled member.
The results of this study highlight the necessity of devising more elaborated methods for calculating a team’s performance-- methods covering different aspects of players' behavior such as skills, strategy, or goals.

\end{abstract}



\section{Introduction}

Online competitive games have become increasingly popular due to the abundance of playing options, continuous updates, and supporting interactions among players.
These games often involve two or more players/teams competing in a match against each other to accomplish a goal.
The goal may include achieving the highest number of points at the end of the match or eliminating the opponents to be the last player or team standing.
One of the main objectives of these games is to ensure balanced matches by matching players based on their skills. 
These games use rating systems for this purpose.

A rating system refers to the algorithmic way of modeling players based on their skills.
These systems use numerical values, referred to as skill ratings, to represent the skill level of players.
The ratings are updated after each match based on the outcome.
Rating systems use ratings to calculate the probability of winning for each side of a match.
These systems then use the calculated probabilities to predict rank and create balanced match-ups.

Rating of players can be aggregated to estimate the rating of a team of players. 
A common approach is to sum the ratings of members.
We refer to this approach as the SUM aggregation method.
This method assumes that each player in the team contributes equally to the total performance of the team.
However, players show different level of performance based on their skills and experience.
For example, top players often play a more important role than amateurs in determining the outcome of a match.

In this paper, we consider three aggregation techniques, including the conventional SUM method along with MAX and MIN.
The MAX method assumes the performance of a team is equal to the maximum of skill ratings of its members, meaning the best player on the team determines the overall performance of the team.
The MIN method assumes the performance of a team is equal to the minimum of skill ratings of its members, meaning the least skilled player on the team determines the overall performance of the team.
While these techniques appear simple, they make dramatically different assumptions about the competitive landscape of online games.
The MIN method assumes that a ``weak link'' can hurt a team, while the MAX method assumes that a strong player can ``carry'' the rest of the team.  These simple methods allow us to infer how the skill differences within a team impacts their chance of success.

We implement these aggregation techniques using three popular rating systems, Elo, Glicko, and TrueSkill, to compare their rank prediction performance.
We perform our experiments on three real-world datasets including PlayerUnknown's Battlegrounds, League of Legends, and Counter Strike: Global Offensive.
We use normalized discounted cumulative gain (NDCG) and accuracy to evaluate rank predictions.
Our results show that rating systems demonstrate better predictive performance when using MAX as their skill aggregation method.

The rest of this paper is organized as follows: In Section~\ref{ref: related}, the related works are reviewed.
In Section~\ref{ref: rating}, we describe rating systems and explain how they predict rank.
In Section~\ref{ref: aggregation}, we detail the skill aggregation methods.
In Section~\ref{ref: method}, the datasets, experimental set-up, and evaluation metrics are explained.
In Section~\ref{ref: results}, we discuss the results in detail.
Finally, we conclude the paper and mention future works in Section~\ref{ref: conclusion}.


\section{Related Work}
\label{ref: related}

Accessibility to high-speed internet and prevalence of streaming platforms such as Twitch have triggered an explosive amount of popularity toward online competitive games.
Many games that were originally introduced as single player offline games have extended their game-play to online platforms while other publishers develop games exclusively for online competition.
The most popular game-play modes of these games include head-to-head and free-for-all games.

In a head-to-head match, two players or teams compete against each other and the winner is typically the one with higher points at the end of the match.
In a free-for-all match, there are more than two sides; several players or teams compete against each other and the winner is the one with the highest number of points at the end of the match (referred to as deathmatch) or the last one standing (referred to as battle royale).
These games use rating systems to match players based on their skills for ensuring balanced matches.

Rating systems simplify the complex concept of skill by performing statistical estimations and using a single number to represent the skill level of players.
For example, 
TrueSkill~\cite{herbrich2007trueskill} assumes that players' skills can be described by the mean of a Gaussian distribution.
Rating systems use estimated skill ratings to predict the outcome of matches.

In a head-to-head game between two players, estimating skills is a straightforward task as the outcome is often directly determined by the performance level of each player.
On the other hand, when two teams compete against each other, their overall performance is highly influenced by the individual performance of their members.
A team of skilled players might lose the game against a team of average players if one of its members fails to keep up with the other members.
This is especially the case for strategy-oriented games where team members are assigned different roles with certain duties.
The role of each team member is even more accentuated when there are multiple teams competing in the same match.
Estimating a team's skill becomes more convoluted as the complexity of the game-play increases.

The majority of rating algorithms sum the performance of members to calculate the overall performance level of a team.
We refer to this method as the SUM approach.
This approach was first introduced by TrueSkill~\cite{herbrich2007trueskill}.
The SUM approach was extensively used by many research works to introduce new rating systems or compare different rating algorithms in shooter games~\cite{zhang2010factor, weng2011bayesian, delalleau2012beyond}, strategy games~\cite{chen2017player, prakannoppakun2016skill, hodge2019win}, and sports~\cite{ibstedt2019application}.
Other works averaged the performance of team members to estimate the team's skill.
This approach was also used in shooter games~\cite{menke2008bradley,delong2011teamskill}, strategy games~\cite{pradhan2020power, aryanata2017prediction}, and sports games~\cite{williams2015abstracting, nikolakaki2020competitive}.
When the competing teams have the same number of players, these two approaches present similar information.
Also, both approaches assume that members equally contribute to the team's performance.
However, the team's performance highly depends on the playing behavior of its members.
A team consisting of one highly skilled player and three amateur players may still achieve a good rank if the skilled player can lead the team or ``carry'' their teammates.
On the contrary, a team consisting of three highly skilled players and one amateur player may not perform well because the amateur player may hinder the playing strategy of the other members, for example, by exposing their positioning to an enemy team.
In these cases, the degree of contribution is different for each member.

In this paper, we focus on teams with equal number of players and thus, only consider the SUM method as the mainstream aggregation technique.
We also include two additional methods; MAX and MIN.
MAX assumes that the performance level of a team is determined by the performance level of its most skilled member.
On the other hand, MIN assumes that the performance of a team diminishes to the performance level of the member with the lowest skill level.
We incorporate these methods into three popular rating systems and compare their predictive performance on two head-to-head games as well as a team battle royale game.
While research often focuses on comparing different skill rating approaches, we aim to evaluate different approaches of team skill aggregation.
We perform our evaluations on three groups of data: all players, best players, and most frequent players.
This way, instead of looking solely at the outcome of matches, we base our comparisons on the frequency of play and playing behavior of players.


\section{Rating Systems}
\label{ref: rating}

Rating systems are one of the main pillars of online competitive games responsible for matching players and teams.
These systems model players' skills using numerical representations.
The numbers are updated after each match based on the results.
Rating systems differ mainly in the way they update those numbers.
In the reminder of this section, we introduce three common algorithms: Elo, Glicko, and TrueSkill.


\subsection{Elo}


Elo~\cite{elo1978rating} has been used as the main rating algorithm to statistically estimate the players' relative skills in many competitive games.
It assumes that a player's skill follows a Gaussian distribution with a mean of $\mu$ and a fixed standard deviation.
The value of $\mu$ is referred to as the skill rating and is used to model the players.





By its original design, Elo is only applicable to head-to-head matches between two players.
We extend Elo to be applicable to any type of match-ups and any number of players or teams.

In a field of \textit{F} where \textit{N} teams compete against each other at the same time, the probability of winning for team $t_i$ can be calculated as:

\small
\begin{equation*}
     \resizebox{.777\hsize}{!}{$Pr(t_i\:wins, F) = \frac{\sum\limits_{1 \leq j \leq N, i \neq j }  \big({1 + e^{\frac{(\mu_{t_j} - \mu_{t_i})}{D}}}\big)^{-1}}{{\binom{N}{2}}}$} 
\end{equation*}
\normalsize

\noindent where $\mu_{t_i}$ and $\mu_{t_j}$ represent the skill rating of team $t_i$ and team $t_j$.
Parameter \textit{D} controls the impact of the difference between ratings.

The observed outcome $R^{obs}_{t_i}$, representing the actual rank of team $t_i$ at the end of the match, is normalized so that the observed ranks sum up to 1:

\small
\begin{equation*}
\label{eqn:R}
    R^{'}_{t_i} = \frac{N - R^{obs}_{t_i}}{\binom{N}{2}}
\end{equation*}
\normalsize

The rating of team $t_i$ can then be updated as:

\small
\begin{equation*}
    \mu_{t_i}^\prime = \mu_{t_i} + K[R^{'}_{t_i} - Pr(t_i\:wins, F)]
\end{equation*}
\normalsize

\noindent where \textit{K} is a weighting factor determining the magnitude of the change to team's rating.

If team $t_i$ consists of players $p_1, p_2, ..., p_n$, the rating of each team member $p_j$ is updated as:

\small
\begin{equation*}
\label{eqn:elo_update2}
    \mu_{j}^\prime = \mu_{j} + w_{p_j}(\mu_{t_i}^\prime - \mu_{t_i})
\end{equation*}
\normalsize

\noindent where $w_{p_j}$ is the contribution weight of player $p_j$ calculated as the ratio of the player's rating to the sum of ratings of their team members.

Elo is often associated with reliability issues resulted from considering a fixed variance for players' skills.
Glicko rating system was one of the first algorithms that addressed this issue.


\subsection{Glicko}
Similar to Elo, Glicko rating system~\cite{glickman1995glicko} assumes that players' skill follows a Gaussian distribution with a mean $\mu$ and standard deviation $\sigma$.
However, Glicko considered $\sigma$ as the rating deviation representing the uncertainties about a player's rating.
The value of $\sigma$ decreases as the player plays more games.
Both $\mu$ and $\sigma$ are updated after each match based on the observed outcome.

While Glicko extended Elo by applying a dynamic skill deviation, like Elo, it is only applicable to head-to-head games between two players.
We extend Glicko to accept any type of match-ups as we did with Elo.

In a field of \textit{F} where \textit{N} teams simultaneously compete against each other, the  probability  of  winning  for team $t_i$ can be calculated as:

\begin{equation*}
    \resizebox{.99\hsize}{!}{$Pr(t_i\:wins, F) = \frac{\sum\limits_{1 \leq j \leq N, i \neq j } \big({1 + 10^{\frac{-g(\sqrt{\sigma_{t_i}^2 + \sigma_{t_j}^2})(\mu_{t_i} - \mu_{t_j})}{D}}}\big)^{-1}}{\binom{N}{2}}$}
\end{equation*}

\noindent where $\mu_{t_i}$, $\mu_{t_j}$, $\sigma_{t_i}$, and $\sigma_{t_j}$ represent the skill ratings and skill deviations of teams $t_i$ and $t_j$.
Parameter \textit{D} is a scaling factor determining the impact of difference between the two ratings.
The function \textit{g} weighs the difference between the two ratings based on their associated deviations. 
It is defined as:

\small
\begin{equation*}
    g(\sigma) = \big(\sqrt{\frac{1 + 3(0.0057565)^2\sigma^2}{\pi^2}}\big)^{-1}
\end{equation*}
\normalsize

Similar to Elo, the observed rank for team $t_i$ is updated to $R^{'}_{t_i}$ so that all observed ranks sum up to 1.
The rating of team $t_i$ can then be updated as:

\small
\begin{equation*}
    \mu_{t_i}^{\prime} = \mu_{t_i} + \frac{0.0057565}{\frac{1}{\sigma_{t_i}^2} + \frac{1}{d_{t_i}^2}} \big[ \sum\limits_{\substack{1 \leq j \leq N \\ i \neq j }}g(\sigma_{t_j})(R^{'}_{t_i} -  Pr(t_i\:wins, F))\big] 
\end{equation*}
\normalsize

\noindent where $d^2$ is minus of inverse of Hessian of the log marginal likelihood and is calculated as:

\small
\begin{equation*}
   \resizebox{\hsize}{!}{ $d^2_{t_i} = \big[ (0.0057565)^2 \sum\limits_{\substack{1 \leq j \leq N \\ i \neq j }}g(\sigma_{t_j})^2 Pr(t_i\:wins, F)(1- Pr(t_i\:wins, F)) \big]^{-1}$}
\end{equation*}
\normalsize

If team $t_i$ includes players $p_1, p_2, ..., p_n$, the skill rating and deviation of player $p_j$ are updated as:

\small
\begin{equation*}
    \mu_{j}^\prime = \mu_{j} + w^{\mu}_{p_j}(\mu_{t_i}^\prime - \mu_{t_i})
,\:\:\:\:\:\:
    \sigma_{j}^\prime = \sigma_{j} + w^{\sigma}_{p_j}(\sigma_{t_i}^\prime - \sigma_{t_i})
\end{equation*}
\normalsize

\noindent where $w^{\mu}_{p_j}$ and $w^{\sigma}_{p_j}$ are the contribution weights of the skill rating and deviation for player $p_j$ calculated as the ratio of the player's skill rating and deviation to the sum of skill ratings and deviations of their team members.

Elo and Glicko were originally designed to work with head-to-head matches between two players.
TrueSkill extended both of these algorithms to be applicable to any type of match-ups.


\subsection{TrueSkill}

Introduced by Microsoft, TrueSkill~\cite{herbrich2007trueskill} is the first official rating system that is not limited to head-to-head games.
It works based on Bayesian inference and leverages a combination of factor graphs and expectation propagation algorithm to rate players' skills.

Similar to Glicko, TrueSkill models players' skills using two values, a mean $\mu$ denoting the skill rating and a standard deviation $\sigma$ representing the uncertainty of the system about the rating.
These values are updated after each match by comparing predicted ranks with observed ranks.
The updating method depends on whether a draw is possible.
For a non-draw case, if $\mu_{t_i}$, $\mu_{t_j}$, $\sigma_{t_i}$, and $\sigma_{t_j}$ represent skill ratings and deviations of teams $t_i$ and $t_j$, assuming team $t_i$ wins the match against team $t_j$, its skill rating is updated by:

\small
\begin{equation*}
    \mu_{t_i}^{\prime} = \mu_{t_i} + \frac{\sigma^{2}_{t_i}}{c}\big[ \frac{N(\frac{t}{c})}{\Phi(\frac{t}{c})}\big]
\end{equation*}
\normalsize

\noindent where $t = \mu_{t_i} - \mu_{t_j}$ and $c = \sqrt{2\beta^2 + \sigma_{t_i}^2 + \sigma_{t_j}^2}$. 
$N$ and $\Phi$ represent the probability density and cumulative distribution functions of a standard normal distribution.
The parameter $\beta$ is the scaling factor determining the magnitude of changes to ratings.
The skill deviation of team $t_i$ is updated by:

\small
\begin{equation*}
    \sigma_{t_i}^{\prime} =  \sigma_{t_i} - \sigma_{t_i} \big(\frac{\sigma_{t_i}^2}{c^2}  \big[\frac{N(\frac{t}{c})}{\Phi(\frac{t}{c})}\big]    \big[\frac{N(\frac{t}{c})}{\Phi(\frac{t}{c})} + t\big] \big)
\end{equation*}
\normalsize

TrueSkill was the first algorithm applicable to team games.
To this end, TrueSkill sums the ratings of team members to calculate the rating of their team.
Most of the ensuing rating systems leveraged a similar approach in their calculations.


\subsection{PreviousRank}

Besides the mainstream rating systems used in our study, we also introduce a naive baseline, PreviousRank.
PreviousRank assumes that players rank similar to what they ranked in their previous match.
If a player is new to the system, we assume that their PreviousRank is equal to $\frac{N}{2}$ where \textit{N} is the number of players competing in the match.
To calculate a team's PreviousRank, we simply add up the PreviousRank of each member.
The team with lower PreviousRank is predicted to win the match.


\section{Aggregation Methods}
\label{ref: aggregation}

In this section, we describe three methods for calculating the skill rating of a team from the ratings of its members.

\subsection{SUM}
SUM is the conventional method used in the majority of skill rating algorithms.
The main assumption of this method is that team members equally contribute to the performance level of the team regardless of their skill level.
If team $t$ consists of \textit{n} players, $p_1, p_2, ..., p_n$, with skill ratings of $\mu_1, \mu_2, ..., \mu_n$, the rating of the team $\mu_t$ is calculated as the sum of ratings of its members:

\small
\begin{equation*}
    \mu_{t} = \sum^{n}_{i=1}{\mu_{i}}
\end{equation*}
\normalsize


\subsection{MAX}
The assumption behind the SUM method holds true when team members have relatively similar skill and experience levels.
However, this assumption is often violated since skill-based matchmaking systems generally place players with different skill and experience levels into a team.
MAX method alleviates this issue by assuming that if a team consists of players with different skill levels, its performance is indicated by the performance of the most skilled member.
If team $t$ consists of \textit{n} players, $p_1, p_2, ..., p_n$, with skill ratings of $\mu_1, \mu_2, ..., \mu_n$, the rating of the team $\mu_t$ is calculated as the maximum of ratings of its members:

\small
\begin{equation*}
    \mu_{t} = \small \operatorname*{arg\,max}_{i \in \{1,2,...,n\}} \normalsize  (\mu_i)
\end{equation*}
\normalsize


\subsection{MIN}
To propose an alternative for addressing the issue of the SUM method, we also include the MIN method.
The main assumption of this method is that if a team consists of players with different skill levels, its performance will reduce to the performance of its least skilled member.
If team $t$ consists of \textit{n} players, $p_1, p_2, ..., p_n$, with skill ratings of $\mu_1, \mu_2, ..., \mu_n$, the rating of the team $\mu_t$ is calculated as the minimum of ratings of its members:

\small
\begin{equation*}
    \mu_{t} = \small \operatorname*{arg\,min}_{i \in \{1,2,...,n\}} \normalsize  (\mu_i)
\end{equation*}
\normalsize


\section{Methodology}
\label{ref: method}

In this section, we first introduce the datasets used to perform our experiments.
We then detail our methodology.
Finally, we describe the metrics used for evaluating the results.


\subsection{Datasets}
We considered two types of team games in this study: head-to-head and battle royale.
In head-to-head matches, two teams compete against each other and the winner is the side that achieves a higher score at the end of the match.
Battle royale matches consist of several teams competing against each other at the same time and the winner is the last team standing at the end of the match, regardless of the scores.


\subsubsection{PlayerUnknown’s Battlegrounds}
PlayerUnknown’s Battlegrounds (PUBG) is one of the most popular battle royale games developed and published by PUBG Corporation.
In PUBG, players parachute onto an island and start scavenging for weapons and equipment.
The players then engage in eliminating each other and the winner is the last player or team staying alive at the end of the match.
PUBG can mainly be played in three modes: squad (teams of four), duo (teams of two), and singletons where every player plays for themselves.
The dataset is publicly available on \href{https://www.kaggle.com/skihikingkevin/pubg-match-deaths}{\textit{Kaggle}}.

In this study, we considered duo matches.
The filtered dataset provides in-game statistics such as distance walked, number of kills, and rank for over 25,000 matches and 825,000 unique players.


\subsubsection{League of Legends}
League of Legends (LOL) is one of the most popular multiplayer online battle arena games developed and published by Riot Games.
The game involves two teams of five players defending their own base while attacking the opponent's base.
The winner is the team who destroys the opponent's Nexus, a structure located in the heart of their base.
The dataset was introduced in~\cite{sapienza2018individual} and is publicly available on the \href{https://doi.org/10.7910/DVN/B0GRWX}{\textit{Harvard Dataverse repository}}.

In this study, we considered a sample of the original dataset. 
The sample dataset includes in-game statistics such as number of kills, gold earned, and rank for over 52,000 head-to-head matches and 324,000 unique players.


\subsubsection{Counter Strike: Global Offensive}
Counter Strike: Global Offensive (CS:GO) is one of the most played shooter games on Steam.
The most common mode of the game consists of two teams of five players, counter-terrorist and terrorist, where terrorists plant a bomb in pre-defined bomb-sites and counter-terrorists attempt to defuse the bomb.
When the bomb is planted, the counter-terrorist team wins the game if they successfully diffuse the bomb and vice versa.
On the other hand, if the bomb is not planted, the team that eliminates all the players of the enemy team is the winner of the match.
The dataset is publicly available on \href{https://www.kaggle.com/mateusdmachado/csgo-professional-matches}{\textit{Kaggle}}.

After pre-processing and merging data from different tables, we created a dataset that includes statistics such as map name and rank for over 26,000 matches and 4,900 unique players.



\begin{table*}
  \centering
  \caption{The average predictive performance of each aggregation method and each rating system for PUBG, LOL, and CS:GO datasets}
\begin{adjustbox}{width=490 pt,center}  \begin{tabular}{l l c c c c c c c c c c c c c}
    \cline{3-15}
    \multirow{2}{*}{} &&  \multicolumn{3}{c}{Elo} & & \multicolumn{3}{c}{Glicko} & & \multicolumn{3}{c}{TrueSkill} & & PreviousRank\\ \cline{3-5}  \cline{7-9} \cline{11-13}
    & & SUM & MAX & MIN & & SUM & MAX & MIN & & SUM & MAX & MIN &  \\
    \hline
    \hline
    \multirow{3}{*}{\vtop{\hbox{\strut PUBG}\hbox{\strut (\%NDCG)}}}  & All Players & 60.2 & \textbf{61.5} & 61.3 & & 59.8 & \textbf{61.5} & 61.3 & & 61.7 & \textbf{62.2} & 61.8 & & 60.8   \\
                          & Best Players & 71.7 & \textbf{73.0} & 69.1 & & 69.6 & \textbf{72.8} & 68.1 & & 69.2 & \textbf{72.8} & 68.4 && 68.4    \\
                          & Frequent Players & 66.7 & \textbf{71.4} & 66.9 & & 67.9 & \textbf{71.6} & 67.9 & & 60.3 & \textbf{62.8} & 60.4 && 59.3 \\
    \hline

    \multirow{3}{*}{\vtop{\hbox{\strut LOL}\hbox{\strut (\%Accuracy)}}}   & All Players & 49.2 & \textbf{50.2} & 49.5 & & 49.3 & \textbf{50.1} & 49.8 & & 49.8 & \textbf{50.4} & 49.3 && 47.6    \\
                          & Best Players & 61.3 & \textbf{76.1} & 51.1 & & 59.1 & \textbf{78.4} & 44.3 & & 60.2 & \textbf{76.1} & 26.1 && 42.1    \\
                          & Frequent Players & 49.2 & \textbf{50.4} & 49.7 & & 49.7 & \textbf{50.3} & 50.2 & & 49.5 & \textbf{50.5} & 49.7 && 48.4 \\
    \hline
    \addlinespace

    \multirow{3}{*}{\vtop{\hbox{\strut CS:GO}\hbox{\strut (\%Accuracy)}}} & All Players & \textbf{64.7} & 64.3 & 60.8 & & \textbf{59.1} & \textbf{59.1} & 56.6 & & \textbf{64.3} & 62.7 & 59.7 && 46.8  \\
                          & Best Players & 54.5 & \textbf{59.1} & 51.8 & & \textbf{57.2} & 55.5 & 50.0 & & 54.5 & \textbf{56.4} & 52.8 && 47.5  \\
                          & Frequent Players & \textbf{64.3} & 63.5 & 60.1 & & 59.2 & \textbf{59.6} & 57.4  & & \textbf{63.6} & 62.5 & 60.1 && 46.6\\
    \hline
    \hline

  \end{tabular}
  \end{adjustbox}

  \label{tab1}
\end{table*}


\subsection{Experimental Setup}

For all datasets, we first sorted the matches by their timestamps.
We then retrieved the list of teams and players along with their corresponding ratings.
Players who appeared in the system for the first time were assigned default ratings, 1500 for Elo and Glicko, and 25 for TrueSkill.
We calculated the ratings of teams based on the ratings of team members using three aggregation methods; SUM, MAX, and MIN.

We then sorted the teams based on their ratings and used the resulted order as the predicted ranks for the match.
Players' ratings were updated after each match by comparing the predicted ranks and observed ranks of their teams and their corresponding contribution weights.
The parameters we used for each rating system include \textit{k = 10} for Elo, \textit{D = 400} for Elo and Glicko, and $\beta$ = 4.16 and $\tau$ = 0.833 for TrueSkill. 
We evaluated the performance of rating systems and aggregation methods using three different set-ups.

The first set-up considers all players in the system regardless of how skilled they are or how many games they played.
This set-up includes players who played very few games.
Capturing the true skill level of these players is often impossible since the rating systems do not observe enough games from them.
Therefore, rank predictions may be hampered for matches where teams consist of one or more of these players.

The next set-up evaluates the predictive performance of rating systems and aggregation methods on the best players in the system.
These players often achieve higher ranks and show consistent playing behavior.
We expect the rating systems to achieve more accurate rank predictions for players with consistent behavior.
To identify these players, we sorted the players based on their most recent skill ratings and selected the top 1000 players who had played more than 10 games.
Since these players had competed in different matches, we performed our evaluations on their first 10 games.

The last set-up focuses on the most frequent players in the system.
Playing more games often results in better skills and more consistent playing behavior.
Therefore, similar to the best players, we expect the rating systems to achieve more accurate rank predictions for frequent players.
To identify the most frequent players, we selected all players who had played more than 100 games.
The predictive performance of rating systems was then evaluated on their first 100 games.



\begin{figure*}
  \centering
\begin{tabular}{ >{\centering\arraybackslash} m{1.4cm} >{\centering\arraybackslash} m{4.9cm} >{\centering\arraybackslash} m{4.9cm} >{\centering\arraybackslash} m{4.9cm} }
\hline
PUBG & Elo & Glicko & TrueSkill\\
\hline

All Players&
{\includegraphics[width=4.38cm]{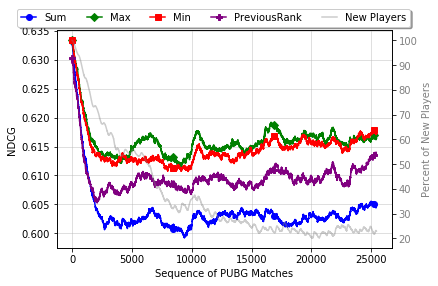}}&
\includegraphics[width=4.38cm]{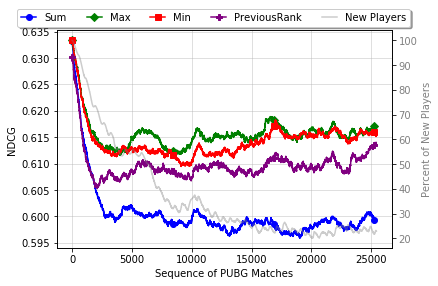}&
\includegraphics[width=4.38cm]{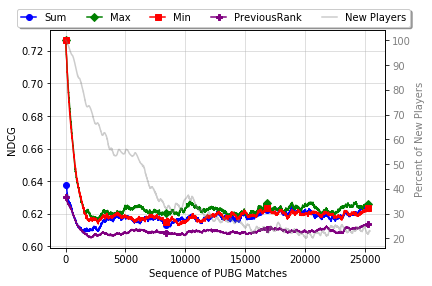}\\

Best Players&
\includegraphics[width=4.0cm]{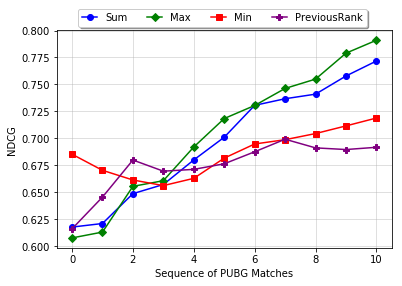}&
\includegraphics[width=4.0cm]{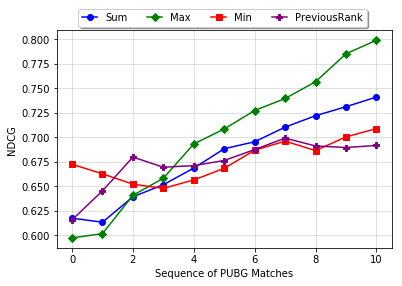}&
\includegraphics[width=4.0cm]{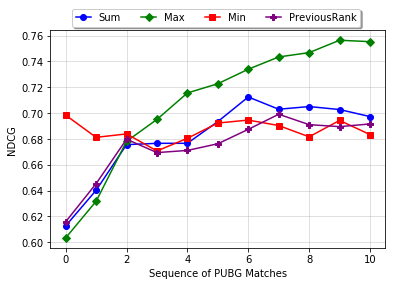}\\

Frequent Players&
\includegraphics[width=4.0cm]{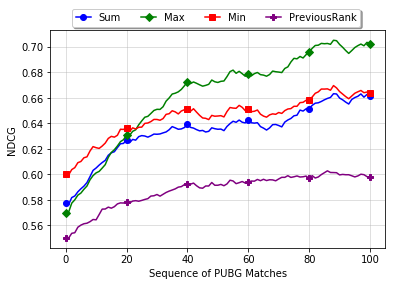}&
\includegraphics[width=4.0cm]{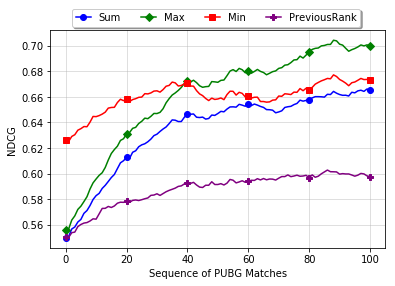}&
\includegraphics[width=4.0cm]{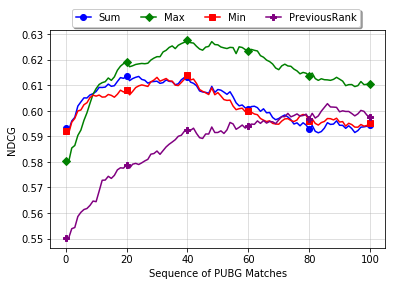}\\
\end{tabular}

\caption{The results of evaluating Elo, Glicko, TrueSkill, and PreviousRank using the SUM, MAX, and MIN aggregation methods in three different experimental set-ups: all players, best players, and frequent players for PUBG.}
\label{fig:pubg} 
\end{figure*}



\begin{figure*}
  \centering
\begin{tabular}{ >{\centering\arraybackslash} m{1.4cm} >{\centering\arraybackslash} m{4.9cm} >{\centering\arraybackslash} m{4.9cm} >{\centering\arraybackslash} m{4.9cm} }
\hline
LOL & Elo & Glicko & TrueSkill\\
\hline

All Players&
{\includegraphics[width=4.38cm]{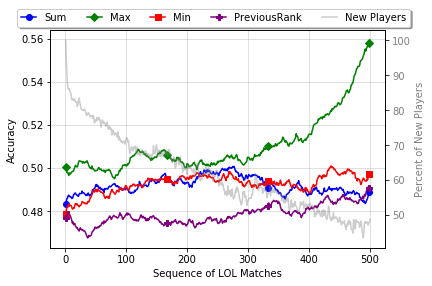}}&
\includegraphics[width=4.38cm]{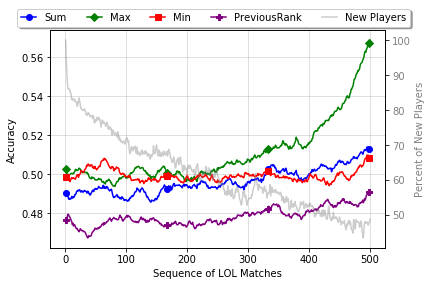}&
\includegraphics[width=4.38cm]{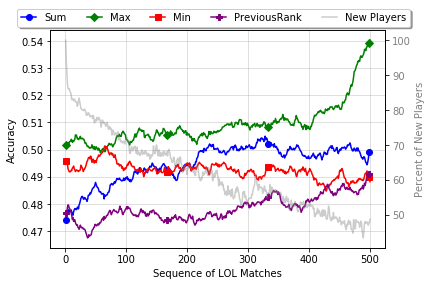}\\

Best Players&
\includegraphics[width=4.0cm]{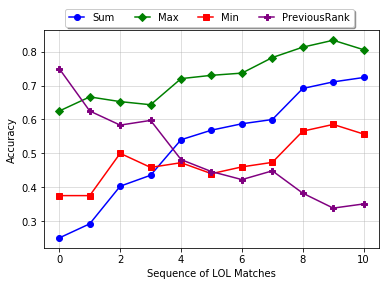}&
\includegraphics[width=4.0cm]{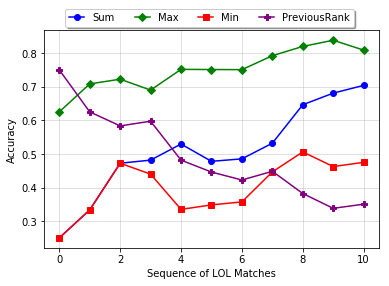}&
\includegraphics[width=4.0cm]{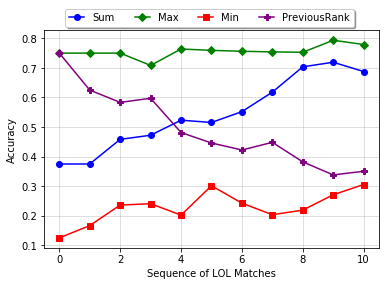}\\

Frequent Players&
\includegraphics[width=4.0cm]{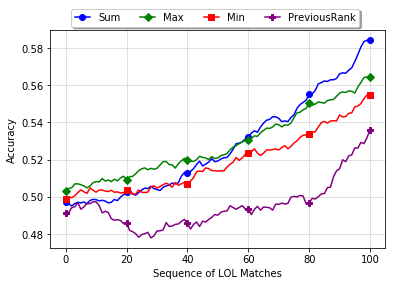}&
\includegraphics[width=4.0cm]{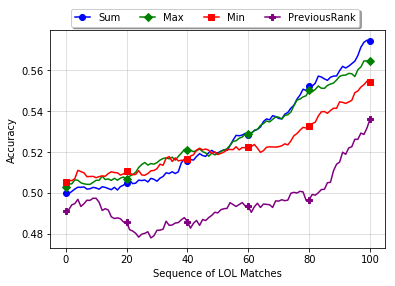}&
\includegraphics[width=4.0cm]{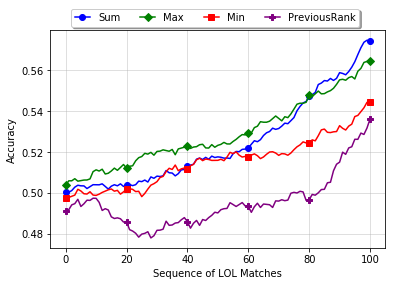}\\
\end{tabular}

\caption{The results of evaluating Elo, Glicko, TrueSkill, and PreviousRank using the SUM, MAX, and MIN aggregation methods in three different experimental set-ups: all players, best players, and frequent players for LOL.}
\label{fig:lol} 
\end{figure*}



\begin{figure*}
  \centering
\begin{tabular}{ >{\centering\arraybackslash} m{1.4cm} >{\centering\arraybackslash} m{4.9cm} >{\centering\arraybackslash} m{4.9cm} >{\centering\arraybackslash} m{4.9cm} }
\hline
CS:GO & Elo & Glicko & TrueSkill\\
\hline

All Players&
{\includegraphics[width=4.38cm]{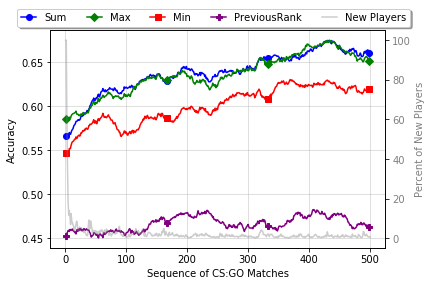}}&
\includegraphics[width=4.38cm]{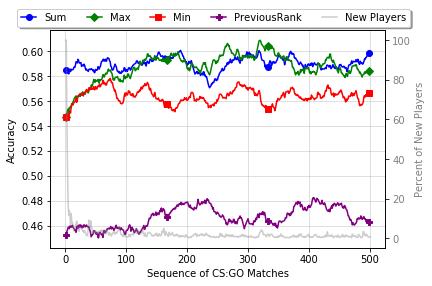}&
\includegraphics[width=4.38cm]{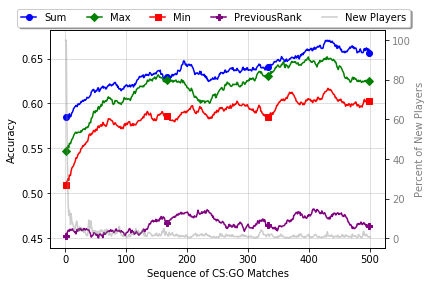}\\

Best Players&
\includegraphics[width=4.0cm]{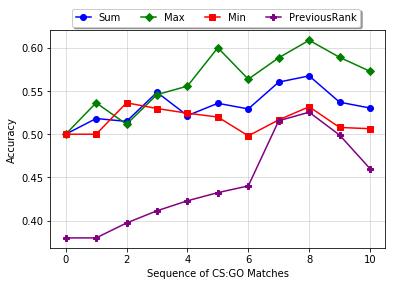}&
\includegraphics[width=4.0cm]{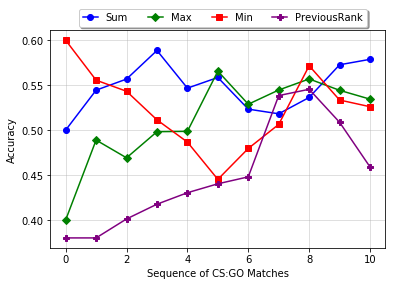}&
\includegraphics[width=4.0cm]{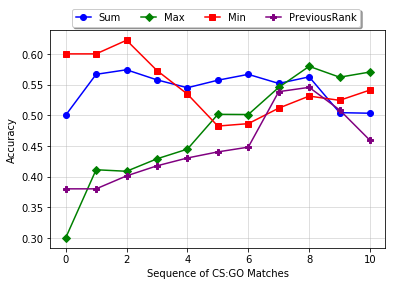}\\

Frequent Players&
\includegraphics[width=4.0cm]{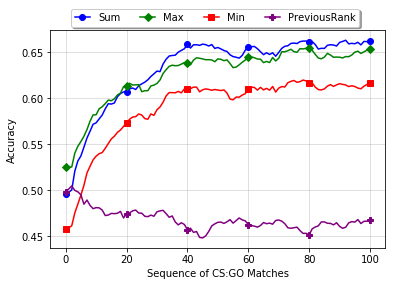}&
\includegraphics[width=4.0cm]{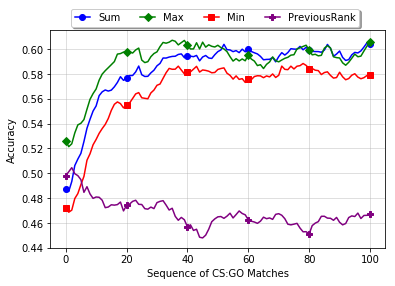}&
\includegraphics[width=4.0cm]{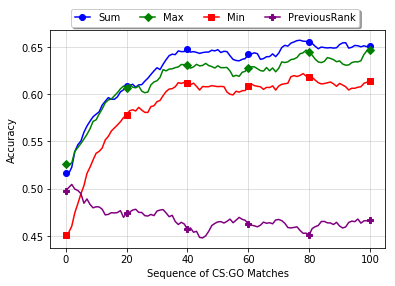}\\
\end{tabular}

\caption{The results of evaluating Elo, Glicko, TrueSkill, and PreviousRank using the SUM, MAX, and MIN aggregation methods in three different experimental set-ups: all players, best players, and frequent players for CS:GO.}
\label{fig:csgo} 
\end{figure*}


\subsection{Evaluation}
Since head-to-head games have only two possible outcomes (three if draw is possible), the resulted ranks can be considered as labels and thus, rank prediction can be thought as rank classification.
Accuracy, a popular metric for evaluating classifications, is a common metric for evaluating the predictive performance of rating systems in these games.
We used accuracy to evaluate predictions for our two head-to-head datasets; LOL and CS:GO.

Battle royale games consist of many teams and players, and the result of a match is often an ordered list of ranks.
Therefore, we used NDCG to evaluate predictions~\cite{dehpanah2020evaluation, dehpanah2021evaluation} for our battle royale dataset; PUBG.



\section{Results and Discussions}
\label{ref: results}

In this section, we present and discuss the results.
Table~\ref{tab1} shows the average scores of aggregation methods for each set-up and each rating system on three datasets.
In this table, rows correspond to experimental set-ups and datasets, and columns correspond to aggregation methods and rating systems.
The best scores are shown in bold.

The results show that rating systems achieve better predictive performance using the MAX aggregation method in the majority of cases.
MAX method outperformed the other two methods in all set-ups for PUBG and LOL, specifically for the best players set-up.
However, for CS:GO, the pattern is not as consistent.
CS:GO has a much higher density compared to the other two datasets, meaning the number of players who played few games is very small and most of the players are known to the system.
Also, we noticed that the skill ratings in this dataset have a significantly smaller range compared to those in the other two datasets.
This causes the ratings of the teams to be closer to each other in comparison, regardless of how a team's overall rating is calculated.

We also evaluated the aggregation methods based on the sequence of matches in the data to compare their performance over time and explore how they adapt to new data.
We expect the models to achieve more accurate skill estimations and rank predictions by observing more games, specifically for players who show consistent playing behavior.
Figures~\ref{fig:pubg}, \ref{fig:lol}, and \ref{fig:csgo} show the results of evaluating aggregation methods in PUBG, LOL, and CS:GO datasets, respectively.
In these figures, results are shown as time series based on the sequence of matches in the datasets sorted by date.
For the all players set-up in LOL and CS:GO datasets, we created the time series by dividing the whole sequence into 500 bins.
Also, a time series of the number of new players in each match is displayed with a gray trend line for the all players set-ups.

Looking at Fig.~\ref{fig:pubg}, the patterns observed for all models in the all players set-up suggest that the MAX method slightly outperforms the MIN method while both of them demonstrate a significantly better performance compared to the SUM method.
The SUM method even shows more inaccurate predictions compared to PreviousRank in the case of Elo and Glicko.
In addition, except for the initial stages of the sequence, all the models demonstrate increasing NDCG values as the number of new players decreases.
However, such a pattern is absent in the case of the SUM method for Glicko.

The results of the best players set-up show all rating systems achieved better prediction performance using the MAX method, especially TrueSkill that uses the SUM method by default.
Both SUM and MAX methods well demonstrated the fact that over time, rating systems achieve a better knowledge of players who show consistent playing behavior and achieve higher ranks.
However, the patterns show that the MIN method struggles to capture the true skill levels of these players.

Similar results can be observed for the most frequent players in this dataset.
The MAX aggregation method considerably outperforms the other methods for all three models.
However, in this set-up, the MIN method shows better performance compared to the SUM method for Elo and Glicko.
The patterns observed in these plots suggest that the methods achieve a better knowledge about the players by observing more games.
The MAX method captures their true skill level significantly faster than the other methods while the SUM and MIN methods show relatively similar rates.

Similar to PUBG, the patterns observed for LOL in Fig.~\ref{fig:lol} for the all players set-up indicate significant superiority of the MAX method over the other two methods.
The SUM and MIN methods show relatively similar performance in this set-up.

The results of evaluating the models for the best players set-up also show more accurate predictions when using the MAX method.
While the SUM method better demonstrates the learning ability of rating systems, the MAX method achieves significantly higher accuracy over the course of ten games.

On the other hand, the difference between the SUM and MAX methods is not as clear for the frequent players set-up.
In this set-up, the MAX method outperforms the other methods for the first 60 to 80 matches after which the SUM method achieves slightly higher accuracy.
The results of this scenario suggest that all aggregation methods demonstrated the learning ability of rating systems from observing more games.
However, such an ability is better manifested in the patterns observed for the MAX and SUM methods.

The results of evaluating aggregation methods for the CS:GO dataset, shown in Fig.~\ref{fig:csgo}, suggest a fairly similar performance for the SUM and MAX methods in the all players set-up.
In most cases, the accuracy of all three methods gradually increases over the sequence of matches as the number of new players in the system decreases.
However, the SUM and MAX methods outperform the MIN method in this scenario.

The patterns observed for the best players set-up differ for each rating system.
For Elo, the MAX method clearly outperforms the other two methods by achieving higher accuracy.
It also better demonstrates the learning ability expected from a rating system for players with consistent playing behavior while the SUM and MIN methods remain almost constant over the whole sequence.
However, the patterns are not as clear for the other two rating systems.
The noticeable point observed in these patterns is that while the SUM and MIN methods show severe cyclical trends, the MAX method almost constantly improves over the sequence of ten matches.
Such patterns are specifically more evident in the case of TrueSkill that uses the SUM aggregation method by default.

Finally, the patterns observed for the frequent players set-up suggest similar performance for the SUM and MAX methods, both significantly outperforming the MIN method.
Also, all incremental accuracy trends indicate that aggregation methods correctly demonstrated the learning ability of rating systems.


\section{Conclusion and Future Work}
\label{ref: conclusion}

In this paper, we examined the effects of different team skill aggregation methods on the predictive performance of rating systems.
We considered the conventional SUM method as well as the MAX and MIN methods.
We performed our experiments on three real-world datasets, two of which consisted of head-to-head games and the other involved battle royale matches.

Our evaluation results showed the relative superiority of the MAX aggregation method.
Rating systems achieved more accurate predictions when using the MAX method in the majority of the tested cases.
This confirms that not all players contribute equally to the performance of their team.
Based on the results, the overall team performance is primarily determined by the performance of the highest skilled member.
Highly skilled players often take the lead in the team and guide their teammates by suggesting efficient strategies through various in-game communication mechanisms such as chat, radio messages, or pings.
In addition, other players often tend to follow such leads as they believe skilled players have higher chances of scoring wins.
On the other hand, if highly skilled players abandon their team and play on their own, there is still a high chance that they outperform many other players in the game and score a high rank for the team anyway.

The results of this study highlighted the necessity of scrutinizing the calculation of a team's performance from individual performance levels of its members.
The two team aggregation alternatives we introduced in this study solely rely on skill ratings computed by traditional rating systems.
As future work, we plan to introduce a comprehensive framework for engineering behavioral features from different sources of data including but not limited to in-game statistics, social network of players, and game logs.
Such features can then be used to model players and teams covering different aspects of game-play such as goals, strategy, and intention besides skills.


\bibliographystyle{IEEETranS}

\end{document}